\documentclass[preprint,12pt,a4paper]{elsarticle}

\usepackage{amsmath}
\usepackage{amssymb}
\usepackage{array}

\usepackage{lineno}

\usepackage{float}
\restylefloat{table}

\usepackage{soul}
\usepackage{enumitem}
\usepackage{subcaption}
\usepackage{rotating}
\usepackage{multirow}
\usepackage[colorlinks]{hyperref}

\hypersetup{breaklinks=true}

\makeatletter
\def\ps@pprintTitle{%
  \let\@oddhead\@empty
  \let\@evenhead\@empty
  \let\@oddfoot\@empty
  \let\@evenfoot\@oddfoot
}
\makeatother

\journal{SoftwareX}

\begin{document}

\begin{frontmatter}



\title{BiometricBlender: Ultra-high dimensional, \\ multi-class synthetic data generator to imitate \\ biometric feature space}

%https://www.elsevier.com/wps/find/journaldescription.cws_home/734734?generatepdf=true


\author[inst1]{Marcell Stippinger}
\author[inst2]{D\'avid Han\'ak}
\author[inst1,inst3]{Marcell T.\ Kurbucz}
\author[inst2]{Gergely Hancz\'ar}
\author[inst2]{Oliv\'er M.\ T\"orteli}
\author[inst1]{Zolt\'an Somogyv\'ari}

\address[inst1]{Wigner Research Centre for Physics, Department of Computational Sciences, 29-33 Konkoly-Thege Mikl\'os Street, H-1121 Budapest, Hungary}
\address[inst2]{Cursor Insight, 20-22 Wenlock Road, N1 7GU London, United Kingdom}
\address[inst3]{Department of Statistics, Corvinus University of Budapest, 8 F\H{o}v\'am Square, H-1093, Hungary}

%\affiliation[inst1]{organization={Wigner Research Centre for Physics, Department of Computational Sciences},%Department and Organization
%            addressline={29-33 Konkoly-Thege Mikl\'os Street}, 
%            city={Budapest},
%            postcode={H-1121}, 
%            country={Hungary}}

%\address{Your institute, some address}
%\affiliation[inst2]{organization={Cursor Insight},%Department and Organization
%            addressline={20-22 Wenlock Road}, 
%            city={London},
%            postcode={N17GU}, 
%            country={United Kingdom}}

\begin{abstract}
  %% Text of abstract, ca. 100 words
  % The purpose of feature screening is to remove irrelevant and redundant features prior to the main analysis. While feature screening is a rapidly growing research field, only a few high or ultra-high dimensional datasets are available publicly that can be employed for benchmarking purposes. In addition, most of them have binary response variables, so they cannot be used in multiple-class cases that are typical in biometric authentication.
  The lack of freely available (real-life or synthetic) high or ultra-high dimensional, multi-class datasets may hamper the rapidly growing research on feature screening, especially in the field of biometrics, where the usage of such datasets is common. This paper reports a Python package called BiometricBlender, which is an ultra-high dimensional, multi-class synthetic data generator to benchmark a wide range of feature screening methods. During the data generation process, the overall usefulness and the intercorrelations of blended features can be controlled by the user, thus the synthetic feature space is able to imitate the key properties of a real biometric dataset.

  % the key properties of a real % \hl{biomedical} or
  % biometric dataset can be imitated by the synthetic feature space.

  % This paper reports a Python package called \hl{BiometricBlender}, which contains an ultra-high dimensional, multi-class data generator for feature screening benchmark. (...) python package for benchmarking feature Screening Benchmark the BiometricBlender python package that is an ultra-high dimensional multi-class Data Generator for feature Screening Benchmark. can be applied for ultra-high dimensional multiclass
\end{abstract}


\end{frontmatter}

%\section*{Required Metadata}
%\label{}

%\section*{Current code version}
%\label{}

%Ancillary data table required for subversion of the codebase. Kindly replace examples in right column with the correct information about your current code, and leave the left column as it is.

\section*{Code and Software Metadata}
\label{sec:metadata}

\begin{table}[H]
\begin{tabular}{|>{\raggedright\arraybackslash}p{1.2cm}|>{\raggedright\arraybackslash}p{5cm}|>{\raggedright\arraybackslash}p{6.2cm}|}
\hline
\textbf{Nr.} & \textbf{Metadata description} & \textbf{Metadata contents} \\
\hline
C1, S1 & Current code version & 1.1.0 \\
\hline
C2, S2 & Permanent link to code/repository used for this code version & \url{https://github.com/cursorinsight/biometricblender/tree/paper} \\
\hline
C3 & Code Ocean compute capsule & 
Not applicable \\
%$https://codeocean.com/2017/07/30/neurospeech-colon-an-open-source-software-for-parkinson-apos-s-speech-analysis/code$\\
\hline
C4, S3 & Legal Code License & MIT License\footnotemark \\
\hline
C5 & Code versioning system used & Git \\
\hline
C6 & Software code languages, tools, and services used & Python \\
\hline
C7,\linebreak S4--S5 & Compilation and installation requirements, operating environments \& dependencies & Python~3.7.1+, h5py~2.10+, numpy~1.18+, scipy~1.6+, scikit-learn~0.24+; OS agnostic (Linux, OS~X, MS~Windows) \\
\hline
C8, S6 & Link to developer documentation and user manual &
\url{https://github.com/cursorinsight/biometricblender/blob/paper/README.md} \\
\hline
C9, S7 & Support email for questions & \texttt{stippinger.marcell@wigner.hu} \\
\hline
\end{tabular}
\caption{Code and software metadata}
%\label{} 
\end{table}
\footnotetext{More information at: \url{https://mit-license.org} (retrieved: 2 December 2021).}

%\linenumbers


%The permanent link to code/repository or the zip archive should include the following requirements: 

%README.txt and LICENSE.txt.

%Source code in a src/ directory, not the root of the repository.

%Tag corresponding with the version of the software that is reviewed.

%Documentation in the repository in a docs/ directory, and/or READMEs, as appropriate.

\section{Motivation and significance}
\label{sec:motivation}

Analyzing ultra-high dimensional data that include hundreds of thousands of features is becoming an increasingly common problem in many fields of modern scientific research \citep{liu2018quantile}. %, such as genomics, biomedical imaging, finance, neurology, and biometrics authentication. %\citep{liu2018quantile}.
Since these datasets typically contain only a relatively few relevant, non-redundant predictors, a screening step that removes irrelevant features prior to the main analysis is often employed for reaching a better prediction accuracy and much faster computation \citep{qiu2020grouped}.

While numerous screening methods have been published in recent years (e.g., \cite{mai2013kolmogorov, mai2015fused, chen2018robust, chen2020model, chen2021efficient, he2019robust, hu2021feature}), only a few high or ultra-high dimensional datasets are available publicly that can be employed for benchmarking purposes. Furthermore, these public datasets (see, e.g, high dimensional datasets related to classification tasks on the UC Irvine Machine Learning Repository\footnote{Available at: \url{https://archive.ics.uci.edu} (retrieved: 2 December 2021).}) typically do not contain ground truth side information on the usefulness of the features. Besides, most of them have binary response variables, so they cannot be used to benchmark methods developed for solving multiple-class screening problems. While in biometrics such problems are typically encountered, it is difficult to imitate the properties of these kinds of feature spaces by using available data generators (e.g., the Madelon dataset \citep{guyon2004result} and the associated data generation algorithm implemented by the \texttt{make\_classification} function of the \textit{scikit-learn} Python package \citep{pedregosa2011scikit}).

% \hl{with an ?improved? implementation in the} \texttt{make\_classification} function of the \textit{scikit-learn} Python package \citep{pedregosa2011scikit}).

To remedy this shortcoming, this paper reports a Python package called BiometricBlender, which is an ultra-high dimensional, multi-class synthetic data generator to benchmark a wide range of feature screening methods. During the data generation process, the overall usefulness and the intercorrelations of features can be controlled by the user. Accordingly, the key properties of a biometric dataset can be imitated by the blended synthetic feature space. This dataset provides an alternative to real biometric datasets, which are typically not freely available. Therefore, it enables the publishing of results achieved on such data.

The paper is organized as follows. \autoref{sec:description} contains the detailed description of the data generator software. As an illustrative example, \autoref{sec:example} presents a synthetic feature space generated to imitate a real-life signature verification dataset. Finally, \autoref{sec:impact} summarizes the impact of the software and provides the conclusions.

\section{Software description}
\label{sec:description}

This section describes the full generator pipeline of BiometricBlender in detail. The output of the pipeline is a high dimensional, multi-class \(S \times F^{visible}\) feature matrix \(\mathbf{V}^{visible} = \left[v^{visible}_{ij}\right]\), where:

\begin{itemize}[noitemsep]
\item \(F^{visible}\) is the desired number of observable, \emph{visible} features;
\item \(\mathcal{F}^{visible} = \left\{f^{visible}_j | 1\leq j \leq F^{visible} \right\}\) is the set of visible features;
\item \(C = |\mathcal{C}|\) is the number of classes;
\item \(S_\mathcal{C}\) is the number of samples per class\footnote{Note that, for simplicity, scalar \(S_\mathcal{C}\) was used. Further development could provide a more realistic feature space by employing classes with different sample sizes.}; and
\item \(S = |\mathcal{S}| = C\cdot S_{\mathcal{C}}\) is the total number of samples.
\end{itemize}

Visible features are derived from a set of \emph{hidden} features, which are significantly fewer than their visible counterparts. In this context:

\begin{itemize}[noitemsep]
\item \(F^{hidden} \ll F^{visible}\) is the desired number of hidden features;
\item \(\mathcal{F}^{hidden} = \left\{f^{hidden}_j | 1 \leq j \leq F^{hidden} \right\}\) is the set of hidden features;
\item \(\mathcal{F}^{true}\) is the set of hidden features which are created to be significant and distinguishing, and
\item \(\mathcal{F}^{fake}\) is the set of hidden features which are just pure noise, and do not contribute useful information to the classification of samples. Moreover:
  \begin{align*}
    \mathcal{F}^{true} \cup \mathcal{F}^{fake} &= \mathcal{F}^{hidden}, \\
    \mathcal{F}^{true} \cap \mathcal{F}^{fake} &= \emptyset.
  \end{align*}
\end{itemize}

With an analogy taken from genetics, hidden features are the \emph{genotypes}, visible features are the \emph{phenotypes} of samples. ``True feature'' genes have an effect on the behavior being observed, while ``fake feature'' genes do not.

If hidden features were directly observable and ideally distributed, sample classification would be a trivial task. The blender components in the second half of the pipeline (see below) ensure that this information is more concealed in the visible features. The full pipeline performs the following steps:

\begin{enumerate}[noitemsep]
\item A suitable distribution type and a set of distribution parameters are selected per \(c_k \in \mathcal{C}\) class and \(f^{hidden}_j \in \mathcal{F}^{hidden}\) hidden feature;
\item \(F^{hidden}\) hidden feature values are drawn from these distributions per sample:
  \[
    \forall s_i \in \mathcal{S}, f^{hidden}_j \in \mathcal{F}^{hidden} : v^{hidden}_{ij} = v(s_i, f^{hidden}_j);
  \]
\item Hidden features are combined with each other through polynomial, linear and/or logarithmic combinations to produce \(F^{visible}\) visible features per sample;
\item A certain amount of random noise is added to the visible feature values.
\end{enumerate}

\subsection{Software Architecture}
\label{ssec:architecture}

\begin{figure}[ht]
  \centering
  \includegraphics[width=\textwidth]{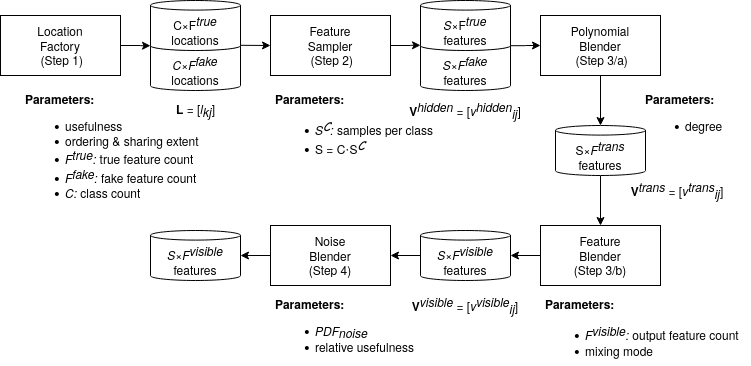}
  \caption{\label{fig:arch} Main components of the generator
    pipeline of BiometricBlender}
\end{figure}

\autoref{fig:arch} shows an overview of the pipeline, with the data flow between the components. The number in parentheses after the name of each component indicates which of the above steps the given component participates in. Each component has a number of parameters that control the usefulness and signal-to-noise ratio of the individual features. The components and their parameters are described in \hyperlink{sssec:location_factory_hyp}{Sections} \ref{sssec:location_factory} to \ref{sssec:noise_blender}.

\subsubsection{Location Factory}
\label{sssec:location_factory}
\begin{figure}[ht]
  \centering
  \includegraphics[width=0.7\textwidth]{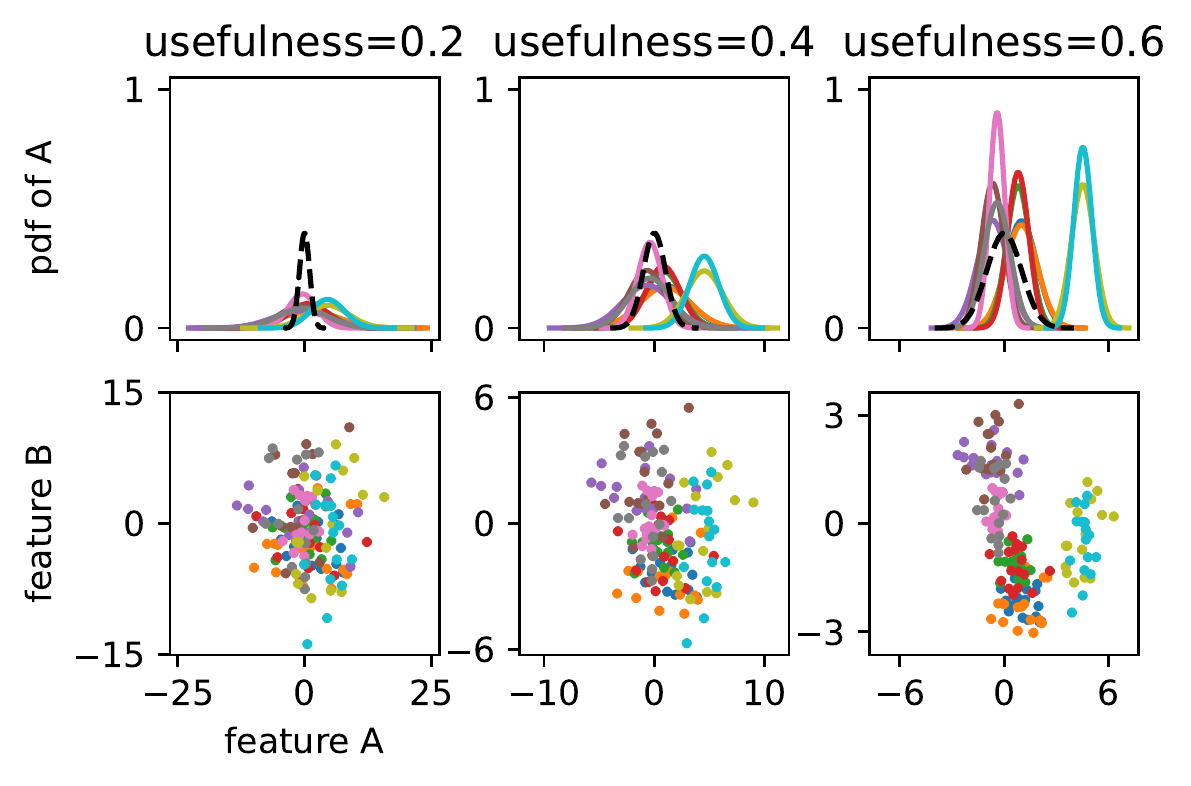}
  \caption{\label{fig:demo} An example of two hidden features A and B for 10 labels with ordering extent 2 and sharing extent 2}
\end{figure}

\hypertarget{sssec:location_factory_hyp}
This component is responsible for determining the \emph{location}\footnote{\emph{Location} defines the translation of a distribution, e.g., it is the mean of normal distributions, and the smallest value, i.e., the start of the range of values for uniform distributions.} of the distributions per class and hidden feature: \(\mathbf{L} = \left[l_{kj} = l(c_k, f^{hidden}_j)\right]\). Locations are randomly chosen under an envelope. % The spread of the envelope controls how easy it is to distinguish individual locations.
%The envelope is a user-defined uniform or normal distribution.
The envelope is either a \emph{normal distribution} or a \emph{uniform distribution}, specified by the user. 
The parameters of the Location Factory are:

\begin{itemize}[noitemsep]
\item \textbf{Number of features} (true and fake): \(F^{true}, F^{fake}\).

\item \textbf{Number of classes:} \(C\).

\item \textbf{Ordering extent} \(\left(\in \mathbb{Z}_{[0, C]}\right)\): controls whether the \(l_{kj} : k \in \mathbb{Z}_{[1, C]}\) sequence of locations of any particular \(f^{hidden}_j\) feature are randomly, partially or fully ordered, thus controls the correlations between features. Its value specifies the average number of locations in every ordered subsequence. This is relevant when not just one but several features come into play: the more ordered locations are, the less detail a new feature adds to the overall amount of information.

Example: \emph{height} and \emph{foot size} are ordered similarly, therefore knowing both does not carry twice as much information as knowing only one of them. \emph{IQ}, on the other hand, is ordered randomly relative to these two features, so knowing both \emph{IQ} and \emph{height} doubles the amount of information.

\item \textbf{Sharing extent} \(\left(\in \mathbb{Z}_{[0, C]}\right)\): controls how many classes share the exact same location on average. With zero sharing extent, all classes have separate, distinguishable locations. With sharing extent \(C\), applied to all fake features, all classes share a single location: \(\forall k \in \mathbb{Z}_{[1, C]} : l_{kj} = l_j\), rendering the feature completely useless. A sharing extent in between creates distinguishable groups of classes, within which groups the individual classes appear identical.

  Example: The sharing extent of \emph{SSN} is zero since all SSNs are unique. The sharing extent of \emph{first names,} on the other hand, is significantly higher.
  
\item \textbf{Usefulness} \(\left(\in \mathbb{R}_{[0, 1]}\right)\) intuitively controls how spread out are the sampling distributions of \emph{Feature Sampler}. The larger the usefulness and the less spread out distributions are, the easier it is to separate feature values generated around these locations. Rather than specifying the usefulness of all hidden features manually, the Location Factory expects a \emph{usefulness scheme}, with which it generates the usefulness of all features. The scheme can be \emph{linear}, \emph{exponential} or \emph{long-tailed}. The usefulness of pure noise features is fixed at zero. \autoref{fig:demo} shows different usefulness parameter settings through the example of two hidden features.

 Example: When identifying people, the usefulness of Social Security Number (SSN) is 1, because it never changes and it is unambiguous. The respiratory rate has much lower, but still non-zero usefulness because while it cannot identify individuals, it can separate some age groups, people doing certain activities or people with some medical condition that affects breathing.
\end{itemize}

\subsubsection{Feature Sampler}
\label{sssec:feature_sampler}

This component takes the \(l_{kj}\) locations and usefulness values of the previous step and draws hidden feature values for the required number of samples from \emph{normal distribution} around these locations. As an option, the \emph{uniform distribution} is also available at the command line interface.
%\emph{uniform or normal distributions} around these locations. 

During sampling, the usefulness is converted to the \emph{scale}\footnote{\emph{Scale} defines the spread of a distribution, e.g., it is the standard deviation for normal distributions, and the length of the range of values for uniform distributions.} of the sampling distributions:

\begin{itemize}[noitemsep]
\item for true features, the converted scale is multiplied by a number drawn from a small uniform distribution around 1, in order to add some variance; and
\item for fake features, a fixed scale value is used. %\hl{Note: \texttt{make\_classification} makes correlated samples (instead of simple scales), do we want such a feature?}

\end{itemize}

For every class \(c_k \in \mathcal{C}\), \(S_{\mathcal{C}}\) samples are created, resulting in a \(S \times F^{hidden}\) matrix \(\mathbf{V}^{hidden} = \left[v^{hidden}_{ij}\right]\) of hidden feature values. Due to the conversion of usefulness to scale, less useful features have larger magnitudes. The magnitudes get normalized just before blending.

\subsubsection{Polynomial Blender}
\label{sssec:polynomial_blender}

This component takes all possible combinations of at least one, at most \(d\) non-unique hidden features, and multiplies them together while preserving their scales by taking the appropriate roots. For example, if feature values are \(x, y \mbox{ and } z\), and \(d=2\), then the generated features are \(x\), \(y\), \(z\), \(\sqrt{\mathstrut xy}\), \(\sqrt{\mathstrut xz}\), \(\sqrt{\mathstrut yz}\), \(\sqrt{\mathstrut x^2}\), \(\sqrt{\mathstrut y^2}\) and \(\sqrt{\mathstrut z^2}\).\footnote{Note that by taking the roots we end up with some repeated values.}

The output is an \(\mathcal{F}^{trans}\) set of \(F^{trans} = |\mathcal{F}^{trans}| = \binom{F^{hidden} + d}{d} - 1 \) (non-unique) transitional features.\footnote{\(F^{trans}\) is equal to the number of combinations of taking exactly \(d\) items of \(F^{hidden} + 1\) items -- all the hidden features plus the constant 1 -- at a time, with replacement; minus the sole case of taking 1 \(d\) times.} The degenerate case of \(d = 1\) results in \(\mathcal{F}^{trans} = \mathcal{F}^{hidden}\).

\subsubsection{Feature Blender}
\label{sssec:feature_blender}

This component takes the transitional features, constructs a random, \({F^{visible} \times F^{trans}}\) dimensional sparse weight matrix \(\mathbf{W} = \left[ w_{ij} \right]\), and produces \(F^{visible}\) blended features using those weights.

The number of blended transitional features per visible feature (i.e., the number of non-zero items in each column of \(\mathbf{W}\)) is randomly chosen from a discrete uniform distribution of small values. The weights themselves are chosen from a \emph{Dirichlet distribution}, such that their sum per visible feature is always 1: \(\forall i : \sum_{j=1}^{F^{trans}}w_{ij} = 1\). Thus the overall magnitude of the features is preserved during blending.

The Feature Blender can operate in two modes:
\begin{itemize}[noitemsep]
\item in \emph{linear} mode, the visible features are weighted sums of the transitional features:
  \[
    \mathbf{V}^{visible} = \left[ v^{visible}_{ij} \right] = \mathbf{V}^{trans} \mathbf{W}^\intercal \text{\enspace where \enspace} v^{visible}_{ij} = \sum_{t = 1}^{F^{trans}} v^{trans}_{it} w_{jt};
  \]
\item in \emph{logarithmic} mode, the visible features are products of the weighted powers of the transitional features:
  \[
    \mathbf{V}^{visible} = \left[ v^{visible}_{ij} \right] \text{\enspace where \enspace} v^{visible}_{ij} = \prod_{t = 1}^{F^{trans}} \left(v^{trans}_{it}\right)^{w_{jt}}.
  \]
\end{itemize}

The linear mode results in feature distributions close to the Gaussian, while logarithmic mode generates long-tailed feature distributions close to the lognormal distribution.\footnote{The linear blending mode (optionally with the noise blending, see below) conforms the generative model of the \emph{Factor Analysis} (FA), thus, in theory, the hidden features may be reconstructed up to a multidimensional rotation and some noise. The logarithmic blending mode is different but the FA still produced reasonable reconstructions.}

\subsubsection{Noise Blender}
\label{sssec:noise_blender}

Finally, optional random noise is added to the visible features, taking the following steps per feature:

\begin{enumerate}[noitemsep]
\item noise is drawn randomly from a normal distribution;
\item an \(\alpha\) relative usefulness is drawn randomly from a uniform distribution between \(0-1\);
\item the feature and the random noise are blended with either linear or logarithmic interpolation, with \(\alpha, 1 - \alpha\) weights, respectively.
\end{enumerate}

\noindent
Thus, if \(\alpha\) is 1, zero noise is added, and when \(\alpha\) is 0, the feature values are completely blocked out by the random noise.

\subsection{Software Functionalities}
\label{ssec:functionalities}

The sole functionality of BiometricBlender is to generate an ultra-high dimensional, multi-class dataset to benchmark a wide range of feature screening methods. The output is generated as an HDF5 file.\footnote{Detailed information about the HDF5 format can be found at: \url{https://hdfgroup.org/solutions/hdf5} (retrieved: 2 December 2021).} Given a fixed seed, the output is reproducible up to rounding errors.

%\subsection{Sample code snippets analysis (optional)}
%\label{ssec:code_snippets}

\section{Illustrative Examples}
\label{sec:example}

As an illustrative example, a synthetic dataset is generated to imitate the private signature feature space of Cursor Insight.\footnote{Cursor Insight won the ICDAR competition on signature verification and writer identification in 2015 \citep{malik2015icdar2015}. For further information, see: \url{https://cursorinsight.com/e-signatures.html} (retrieved: 2 December 2021). Note that, to demonstrate the potential in screening, the dataset generated here is somewhat noisier than the imitated data.}

\begin{samepage}
  The following custom command line parameters were set:
  \begin{itemize}[noitemsep]
  \item \texttt{n-labels} = 100;
  \item \texttt{n-samples-per-label} = 16;
  \item \texttt{n-true-features} = 40;
  \item \texttt{n-fake-features} = 160;
  \item \texttt{average-consecutive-locations} = 2;
  \item \texttt{average-shared-locations} = 3;
  \item \texttt{n-features-out} = 10\,000;
  \item \texttt{blending-mode} = \texttt{'logarithmic'}.
  \end{itemize}
\end{samepage}

\noindent
The resulting dataset has 1\,600 samples and 10\,000 features. Note that in this feature set one must adjust the parameters of the generative model to approximate the statistics (e.g., eigenspectrum) of the output rather than prescribing the statistics themselves. We tested it for classification in the following ways. We trained the \textit{scikit-learn} \citep{pedregosa2011scikit} (version: 0.24.2) implementation
%\footnote{\href{https://scikit-learn.org/0.24/}{Scikit-learn Python package} version 0.24.2}
of three basic classifiers on the original data and on the reduced/decomposed version of the data. We characterized the best cross-validated accuracy that can be attained for each classifier using a full grid search over crucial parameters. These parameters were
\begin{itemize}[noitemsep]
\item\texttt{weights} = \texttt{'uniform'}, \texttt{'distance'} for $k$-nearest neighbors ($k$NN);
\item\texttt{C} = 0.5, 1.0, 2.0; \texttt{tol} = 1e-4, 1e-3, 1e-2 for the support vector classifier (SVC); and
\item\texttt{n\_estimators} = 1000; \texttt{min\_samples\_leaf} = 1, 2, 4; \texttt{max\_depth} = \texttt{None}, 8, 10; \texttt{min\_impurity\_decrease} = 0.0, 0.01, 0.05 for the Random Forest Classifier (RF).
\end{itemize}
\noindent
The reduction step allowed the classifiers to work on a more focused dataset. We executed each reduction/decomposition algorithm to produce a reduced feature space of 10, 25, 50, 100, 200, 400, and 800 features and reported the best accuracy only, see \autoref{tab:accuracy}. The Principal Component Analysis (PCA) kept its default settings. To Factor Analysis (FA) we applied the \texttt{varimax} rotation. The $k$-best \texttt{SelectKBest} method increasingly selected the best features using the \texttt{f\_classif} score. \(\mathcal{F}^{true}\) used the true hidden features.

\begin{table}[!htb]

    % \begin{subtable}{.5\linewidth}
    %   \centering
    %     \caption{Classification performance}
    %     \resizebox{0.96\textwidth}{!}{%
    %     \begin{tabular}{llrrrr}
    %     \hline
    %     \hline
    %     \multicolumn{2}{c}{Reduction} & \multicolumn{1}{c}{None} &    \multicolumn{1}{c}{PCA} &     \multicolumn{1}{c}{FA} &   \multicolumn{1}{c}{$k$-best} \\ \hline
    %     \multirow{3}{*}{\rotatebox{90}{Class.}} & $k$NN        &  0.021 &  0.029 &  0.030 &  0.553 \\
    %     & SVC        &  0.074 &  0.086 &  0.113 &  0.658 \\
    %     & RF         &  0.531 &  0.051 &  0.628 &  0.763 \\
    %     \hline
    %     \hline
    %     \end{tabular}
    %     }
    % \end{subtable}%
    % \begin{subtable}{.5\linewidth}
    %   \centering
    %     \caption{Fit time}
    %     %(b) Fit time
    %     \resizebox{\textwidth}{!}{%
    %     \begin{tabular}{llrrrr}
    %     \hline
    %     \hline
    %     \multicolumn{2}{c}{Reduction} & \multicolumn{1}{c}{None} &    \multicolumn{1}{c}{PCA} &     \multicolumn{1}{c}{FA} &   \multicolumn{1}{c}{$k$-best} \\ \hline
    %     \multirow{3}{*}{\rotatebox{90}{Class.}} & $k$NN      &    0.14s &   0.003s &   0.004s &   0.001s \\
    %     & SVC        &   11s &   0.42s &   0.41s &   0.34s \\
    %     &    RF         &  255s &  49s &  30s &  30s \\
    %     \hline
    %     \hline
    %     \end{tabular}
    %     }
    % \end{subtable}

    \begin{subtable}{.52\linewidth}
      \centering
        \caption{Classification performance}
        \resizebox{\textwidth}{!}{%
        \begin{tabular}{llrrrrr}
        \hline
        \hline
        \multicolumn{2}{c}{Reduction} & \multicolumn{1}{c}{None} &    \multicolumn{1}{c}{PCA} &     \multicolumn{1}{c}{FA} &   \multicolumn{1}{c}{$k$-best}  & \multicolumn{1}{c}{\(\mathcal{F}^{true}\)} \\ \hline
        \multirow{3}{*}{\rotatebox{90}{Class.}} %
        % class      &  None  &  PCA   &  FA    &    k-best   & true
        & $k$NN      &  0.131 &  0.218 &  0.214 &     0.641   & 0.632 \\
        & SVC        &  0.471 &  0.466 &  0.548 &     0.686   & 0.656 \\
        & RF         &  0.609 &  0.371 &  0.716 &     0.692   & 0.860 \\
        \hline
        \hline
        \end{tabular}
        }
    \end{subtable}%
    \begin{subtable}{.47\linewidth}
      \centering
        \caption{Fit time of the classifier}
        %(b) Fit time, reduction time
        \resizebox{\textwidth}{!}{%
        \begin{tabular}{llrrrr}
        \hline
        \hline
        \multicolumn{2}{c}{Reduction} & \multicolumn{1}{c}{None} &    \multicolumn{1}{c}{PCA} &     \multicolumn{1}{c}{FA} &   \multicolumn{1}{c}{$k$-best} \\ \hline
        \multirow{3}{*}{\rotatebox{90}{Class.}}  %
        % class      &  None  &  PCA   &  FA    &    k-best
        & $k$NN      &0.153s &  0.003s & 0.001s &   0.006s \\
        & SVC        &   24s &   0.37s &  0.42s &   0.46s \\
        & RF         &  300s &     22s &    21s &     29s \\ 
        \hline
        \hline
        \end{tabular}
        }
    \end{subtable}
\caption{Classification results on the 1\,600$\times$10\,000 dataset for three basic classifiers and various reduction algorithms. \emph{(a)} Only the best accuracy among all parameters is reported. \emph{(b)} Fit times are the wall time after the reduction step and correspond to the accuracy shown above. }
\label{tab:accuracy}
\end{table}

\section{Impact and conclusion}
\label{sec:impact}
%The main value of
%This paper reports a Python package called \hl{BiometricBlender}, which is an ultra-high dimensional, multi-class data generator to benchmark a wide range of feature screening methods.
The ultra-high dimensional, multi-class data generator called BiometricBlender supports the rapidly growing research on feature screening in two ways. On the one hand, it facilitates the benchmark of a
wide range of feature screening methods (see \autoref{tab:accuracy}) by providing an alternative to real (typically non-free) %biomedical or
biometric datasets.
%, which are typically not freely available.
On the other hand, it enables the publishing of %it makes it possible to publish
results achieved on such data. To this end, the overall usefulness and the intercorrelations of blended features can be controlled by the user during data generation. %,
Thus, the synthetic feature space is able to imitate the key properties of a real %biomedical or
biometric dataset.

\section*{Declaration of Competing Interest}

We wish to draw the attention of the reader to the following facts, which may be considered as potential conflicts of interest, and to significant financial contributions to this work. The nature of potential conflict of interest is described below: some of the authors work for Cursor Insight, an IT company targeting human motion analysis, person classification and identification based on large-scale biometric data in particular. In order to handle such real-life, multi-class, ultra-high dimension datasets efficiently, we came up with our own feature screening algorithm, because we found industry standard solutions insufficient. We have then decided to share our solution with the general public. The demand for a synthetic data generator arose when, in order to prove the performance of our screening algorithm against standard solutions, we started looking for publicly available reference datasets of such dimensions, or generators of such, and found none.

\section*{Acknowledgments}
The authors would like to thank Erika Griechisch (Cursor Insight, London) and Andr\'as Telcs (Wigner Research Centre for Physics, Budapest) for their valuable comments and advice. M.S., M.T.K., and Z.S. thank the support of E\"otv\"os Lor\'and Research Network. Z.S. was supported by the Hungarian National Research, Development and Innovation Office, under grant number NKFIH K135837 and the Hungarian National Brain Research Program 2017-1.2.1-NKP-2017-00002.

%The authors would like to thank Erika Griechisch (Cursor Insight, London) for her valuable comments and advice. M.S., M.T.K. and Z.S. thank the support of E\"otv\"os Lor\'and Research Network. Z.S. was supported by the Hungarian National Research, Development and Innovation Office, under grant number NKFIH K135837 and the Hungarian National Brain Research Program 2017-1.2.1-NKP-2017-00002.


%% %\label{}

%% References:


%\begin{thebibliography}{00}

\bibliographystyle{elsarticle-num}

\bibliography{bibliography.bib}

%\end{thebibliography}
%Please add the reference to the software repository if DOI for software is available. 

%\section*{Current executable software version}
%\label{}

%Ancillary data table required for sub version of the executable software: (x.1, x.2 etc.) kindly replace examples in right column with the correct information about your executables, and leave the left column as it is.

%\begin{table}[!h]
%\begin{tabular}{|l|p{6.5cm}|p{6.5cm}|}
%\hline
%\textbf{Nr.} & \textbf{(Executable) software metadata description} & \textbf{Please fill in this column} \\
%\hline
%S1 & Current software version & For example 1.1, 2.4 etc. \\
%\hline
%S2 & Permanent link to executables of this version  & For example: $https://github.com/combogenomics/$ $DuctApe/releases/tag/DuctApe-0.16.4$ \\
%\hline
%S3 & Legal Software License & List one of the approved licenses \\
%\hline
%S4 & Computing platforms/Operating Systems & For example Android, BSD, iOS, Linux, OS X, Microsoft Windows, Unix-like , IBM z/OS, distributed/web based etc. \\
%\hline
%S5 & Installation requirements \& dependencies & \\
%\hline
%S6 & If available, link to user manual - if formally published include a reference to the publication in the reference list & For example: $http://mozart.github.io/documentation/$ \\
%\hline
%S7 & Support email for questions & \\
%\hline
%\end{tabular}
%\caption{Software metadata (optional)}
%\label{} 
%\end{table}

\end{document}